\pdfoutput=1
\documentclass[11pt]{article}

\usepackage[]{acl}

\usepackage{times}
\usepackage{latexsym}

\usepackage[T1]{fontenc}
\usepackage[utf8]{inputenc}

\usepackage{microtype}

\usepackage{multicol}
\usepackage{multirow}
\usepackage[ruled]{algorithm2e}
\usepackage{float}
\usepackage{graphicx}
\usepackage{courier}
 \usepackage{amsmath} 
 
\title{On the Role of Context for Discourse Relation Classification in Scientific Writing}

\author{ \\
  CSIRO \\
  Sydney, Australia \\
  \texttt{stephen.wan@csiro.au} \\\And
  \\
  Heidelberg Institute for Theoretical Studies \\
  Heidelberg, Germany \\
  \texttt{firstname.lastname@h-its.org}
  }

\author{
Stephen Wan$^{\clubsuit}$  \quad  Wei Liu$^{\diamondsuit}$ \quad Michael Strube$^{\diamondsuit}$ \\
$^{\clubsuit}$ CSIRO Data61, Australia \\
$^{\diamondsuit}$ Heidelberg Institute for Theoretical Studies, Germany \\
\texttt{stephen.wan@data61.csiro.au} \\
\texttt{\{wei.liu, michael.strube\}@h-its.org} \\
}

\newcommand{\fixme}[1]{{\color{orange} {FIXME}}}

\begin{document}
\maketitle
\begin{abstract}
With the increasing use of generative Artificial Intelligence (AI) methods to support science workflows, we are interested in the use of discourse-level information to find supporting evidence for AI generated scientific claims. A first step towards this objective is to examine the task of inferring discourse structure in scientific writing.

In this work, we present a preliminary investigation of pretrained language model (PLM) and Large Language Model (LLM) approaches for \textit{Discourse Relation Classification} (DRC), focusing on scientific publications, an under-studied genre for this task.  We examine how context can help with the DRC task, with our experiments showing that context, as defined by discourse structure, is generally helpful.  We also present an analysis of which scientific discourse relation types might benefit most from context.

\end{abstract}

\section{Introduction}

% \noindent\emph{AI and Science today}
Recent Artificial Intelligence (AI) advances coupled with the agentic AI approach have seen a burst of activity in the area of ``AI for Science'', the application of AI techniques to help accelerate scientific discovery.  Examples include usage of \textit{Google's Co-scientist} \cite{Penades2025AIEvolution}, OpenAI's \textit{Deep Research}\footnote{https://openai.com/index/introducing-deep-research}, NVidia's foundation models for life sciences\footnote{https://www.nvidia.com/en-au/use-cases/biomolecular-foundation-models-for-discovery-in-life-science}  and the agentic AI platform \textit{Future House}\footnote{https://www.futurehouse.org/}.  Many of these tools offer an AI research assistant that helps complex research information needs, such as question answering and research planning.

% \noindent\emph{Idea of explainable AI: annotating surrounding content to generated answers.}
Within these AI for Science applications, generative AI approaches based on Large Language Models (e.g., \citeauthor{DBLP:journals/corr/abs-2005-14165}, \citeyear{DBLP:journals/corr/abs-2005-14165}) are used to generate answers (novel text) to complex questions, introducing the problem of addressing hallucinations and lack of faithfulness (to source references) \cite{fang-etal-2024-understanding}.  

A popular approach to these problems is to show passages from the source material that supports the generated answer.  This approach, sometimes referred to as ``contextualising scientific claims'', was the focus of the \textit{Context24} shared task \cite{Chan2024OverviewClaims}.
Interestingly, the leading contribution in the Context24 shared task demonstrated the utility of scientific discourse cues for detecting such justification material \cite{Bolucu2024CSIRO-LTLiterature}.\footnote{Discourse cues are described as \textit{common expressions} in the original paper.}  
This raises an interesting question: \textit{can discourse information be further employed to help in providing supporting evidence for generative AI answers to scientific questions?}  A necessary precursor to such an approach would be the ability to infer the discourse structure of a given paper.
As a first step towards a study of this topic, in this paper, we focus here on studying the technical challenge of inferring the discourse relations between passages of scientific writing.

\begin{figure}
    \centering
    \includegraphics[scale=0.650]{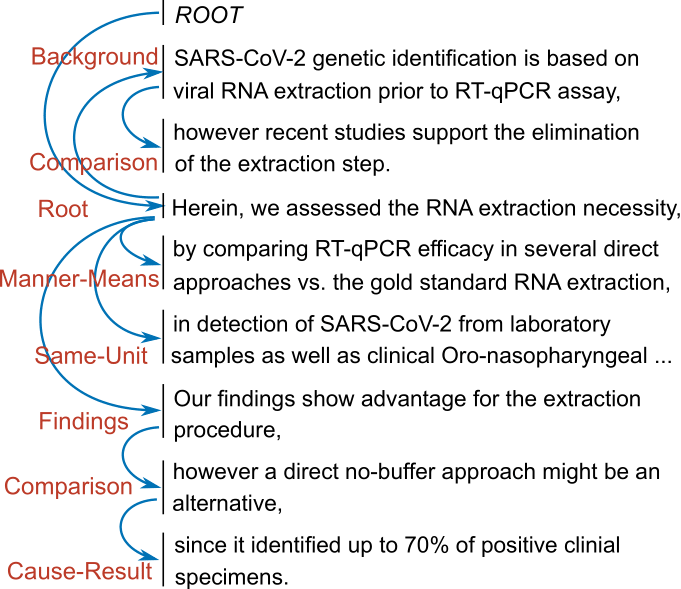}
    \caption{An example of a dependency discourse structure for an abstract for COVID19-related science article.  Figure from \cite{nishida2021outofdomain} }
    \label{fig:dds}
\end{figure}

We focus on data from two discourse datasets for scientific text, SciDTB \cite{yang-li-2018-scidtb} and CovDTB \cite{nishida2021outofdomain}.  Both datasets follow the approach that annotates discourse structure as dependency trees, introduced in the SciDTB approach \cite{yang-li-2018-scidtb}.  To our knowledge, these are the largest discourse datasets currently available.  An example of such a discourse tree is presented in Figure \ref{fig:dds}.

We present an example from the SciDTB dataset \cite{yang-li-2018-scidtb} in Figure \ref{fig:motivation_example}.\footnote{SciDTB dataset, document ID:P14-1131.}  The figure shows segmented elementary discourse units (EDUs) for the arguments of the relation.  The ground truth relation between the two arguments was annotated as \textit{condition}.\footnote{That is, a conditionality for a given situation.}  This classification task is highly ambiguous.  We note that, in this dataset, the word "without" that begins \textit{arg 2} is also associated with two other relations such as \textit{contrast} or \textit{manner}.  This ambiguity can be alleviated with additional context. In the example, \textit{arg 1} is annotated as being preceded by the fragment that has an \textit{elaboration} relation and is about \textit{efficiency}.  Of the possible relations, this context potentially provides additional information that supports \textit{conditionality} as a more favourable interpretation compared to \textit{contrast} or \textit{manner}.\footnote{That is, in the example, \textit{computation} is described as \textit{efficient}, not because of an explicit comparison (contrast) nor an indication of how to perform a task (manner) but by virtue of conditions, in this case, \textit{without} the described \textit{expenses}.}  

\begin{figure}[]
\small
    \centering
    \begin{tabular}{|l|l|}
    \hline
          \multirow{1}{*}{\textit{EDU 1}} & because it can compute a single node similarity \\ 
          \multirow{1}{*}{\textit{EDU 2}} & (rel:condition) \textit{without} having to compute the \\
                                        & similarities of the entire graph . \\
    \hline
    \hline
          \textit{context} & (rel:elaboration)  that is efficient \ldots \\
    \hline
    \end{tabular}
    \caption{An example of how contextual information can help disambiguate which discourse relation holds between two text fragments.}
    \label{fig:motivation_example}
\end{figure}

% \noident\emph{What we want to study is how much information to present (and how)}
% \noident\emph{Compare LLM approaches to scientific DRC against finetuned PLMs}
Inspired by examples such as this and also by recent work in discourse analytics that examines the role of context for DRC in other text genres (e.g., \citeauthor{zhang-etal-2021-context}, \citeyear{zhang-etal-2021-context}; see Section \ref{sec:related} for a comprehensive survey), we are further interested in examining the role discourse context for the DRC task.  As the majority of these studies have focused on the Penn Discourse Treebanks (PDTB) \cite{Webber2019TheManual} and given that scientific writing  (and the discourse relations therein) as notably different from other genres (e.g., see \citeauthor{shi-demberg-2019-next}, \citeyear{shi-demberg-2019-next}), our goal is to study how context affects DRC for scientific writing.
% Specifically, our interest lies in evaluating recent LLMs and Pretrained Language Models (PLMs)  for DRC, examining the role of context for this inference.  
% 
In particular, we are interested in examining whether context selection informed by discourse structure, which we refer to here as \textit{discourse structured context},  has advantages over other methods, such as adjacent text spans.

This paper makes several unique contributions.  We show that structured context helps to improve DRC for scientific writing, as represented by the two datasets, finding that this approach benefits both pre-trained language model and large language model approaches.  Furthermore, we present an error analysis to explore the situations context in which context is helpful, revealing some interesting correspondences between scientific discourse relations within the two datasets.

\section{Related Work}
\label{sec:related}

\subsection{Discourse Relation Classification}

Discourse Relation Classification (DRC), the task of inferring the relation type that holds between two EDUs, has a rich history and is an actively researched area \cite{pitler-etal-2009-automatic,gessler-etal-2021-discodisco,long-webber-2022-facilitating,zhou-etal-2022-prompt-based,liu-strube-2023-annotation}.  The DRC task is typically divided into two types, \textit{explicit} and \textit{implicit} \cite{pitler-etal-2009-automatic}.  In the former, the two text spans are connected by a discourse relation signalled by an observable discourse connective.  The latter is termed implicit as there is an absence of such a connective.  The implicit variant  is regarded  as much harder than the explicit variant \cite{pitler-etal-2009-automatic}.  

In recent years, to facilitate DRC research, shared tasks have been organised within the DISRPT series \cite{zeldes-etal-2021-disrpt, braud-etal-2023-disrpt}.  These shared tasks had the benefit of broadening the datasets used to evaluate DRC approaches, which has tended to focus on the PDTB \cite{Prasad2008The2.0}.  Notably, the two datasets used in this paper were introduced in the 2023 shared task \cite{braud-etal-2023-disrpt}, however, the submitted approaches at the time did not focus on analysing the use of context for these datasets.

\subsection{Context in DRC}
The position survey article of \citet{Atwell2021WhereRecognition} notes that ``most shallow discourse parsers use only the argument pairs
to determine the discourse sense without considering ... context.''  There have, however, been some exceptions. 
The effect of using discursive context on DRC has been studied in the context of annotation quality and annotator confidence in \cite{atwell-etal-2022-role}.  This work examines the role of context in the PDTB 2 \cite{Prasad2008The2.0} and 3 \cite{Webber2019TheManual} datasets and shows that annotation quality improves with context (in this case, preceding text), particularly for certain relationships. In that same work, \citeauthor{atwell-etal-2022-role} translate the insights from human annotations and context into modelling insights, exploring the use of the XLNet model for classification with context, that incorporates some modelling of when context is needed. \citeauthor{atwell-etal-2022-role} note the prior work of \citet{scholman-demberg-2017-crowdsourcing} in examining the linkage between pronoun use and the need for context within PDTB, again from the perspective of acquiring human annotations.

The sequential modelling of adjacent text spans (and their relationships) has been studied by \citet{dai-huang-2018-improving}, \citet{shi-demberg-2019-next}  and \citet{zhang-etal-2021-context}.  These works 
% , whether sentences or paragraphs, 
generally evaluate on PDTB data.  The exception is \citet{shi-demberg-2019-next}, who also evaluate on biomedical data, albeit by mapping biomedical relations to PDTB discourse relations.  \citeauthor{shi-demberg-2019-next} also argue that the Next Sentence Prediction (NSP) capability of the the BERT model is particularly useful for modelling discourse relations.  This same approach is used by \citet{gessler-etal-2021-discodisco}, who also use BERT specifically for its NSP capability but add features relating to the surrounding context.  This included direction features and embeddings of the surrounding sentences (to the text spans being considered).  
% This work differs from the other work in that evaluations are based on the DISRPT 2023 dataset, which includes a broader (and multilingual) range of discourse corpora in addition to data from the PDTB.
% 
\citet{zhang-etal-2021-context} model discourse structure as a graph and use graph representations in a neural network to capture discourse context, showing benefits for discourse relation classification.  These investigations provide alternative representations of context to our study, which does not use specific features or graph representations of context.  Our study differs in that we use various methods to select context, including a consideration of the discourse dependency tree, and we prepend contextual text to the EDUs being judged.

While prior treatments of context have shown that it is useful for DRC, these studies generally focus on non-scientific writing, like the PDTB.  We note that the approach of \citet{gessler-etal-2021-discodisco}, and the following work of \citet{metheniti-etal-2024-feature}, is evaluated on the DISRPT 2021 dataset.  However, while this dataset includes data a range of genres, it does not include scientific articles.  Indeed, \citet{shi-demberg-2019-next} note that the discourse relations are notably different in science literature compared to relations from the PDTB.  We argue that this difference thus merits a dedicated study of the effects of context for science literature.

\subsection{Other Discourse-level Analytics for Scientific Writing}

Related to the task of inferring discourse relations for scientific writing are the tasks of argument zoning \cite{ teufel-etal-2009-towards}, citation function classification and classification \cite{teufel-etal-2006-automatic, Wan2009WhettingContext}, scientific argument mining (e.g., \cite{lawrence-reed-2019-argument, Accuosto2020MiningEmbeddings, binder-etal-2022-full, fergadis-etal-2021-argumentation}  and science communication sentence classification \cite{louis-nenkova-2013-makes, august-etal-2020-writing}; all of which infer a discourse-level features relating to argumentation or scientific writing structure.  For this body of work, we note that the use of context has been studied has been studied within the topic of sentence-level categorisation for scientific function \cite{kiepura-etal-2024-scipara}.  While these associated fields can inform our work, in this paper, we focus on discourse relations rather than argument zones, single sentence classification, or discourse-level relationships across citations.

\section{Data}
\label{sec:data}

Here, we focus on data from prior work in discourse analysis that provides ground truth annnotations for scientific discourse structure, namely the the Covid (Discourse) Dependency Treebank (CovDTB) \cite{nishida2021outofdomain}, and the Science (Discourse) Dependency Treebank (SciDTB) \cite{yang-li-2018-scidtb}.

\begin{table}[]
\small
    \centering
    \begin{tabular}{|c|c|c|}
    \hline
         Dataset &  Train:Dev:Test size & Genres  \\
    \hline
   
         % \multirow{1}{*}{GUM} & \multirow{1}{*}{RST} & \multirow{1}{*}{19497:2618:2576} & News, Travel, Instructions, Science, Biographies, Fiction, Textbooks \\
         % \hline
         CovDTB & (2400):2400:2587 & Science (Biomed) \\
         SciDTB & 6061:1935:1912 & Science (NLP) \\
    
%    \hline
%         PDTB 2 & PDTB & 12632:1183:1046 & News \\
%         PDTB 3 & PDTB & 43920:1674:2257 & News \\
    \hline
    \end{tabular}
    \caption{Overview of the scientific discourse datasets studied in this work.}
    \label{tab:datasets}
\end{table}

\subsection{Discourse Dependency Representations}

The SciDTB and CovDTB datasets use the dependency discourse structure (DDS) introduced in the SciDTB work \cite{yang-li-2018-scidtb}.  Here, structures are directed acyclic graphs, specifically trees.
An example of a DDS is shown in Figure \ref{fig:dds}.  In this structure, nodes are EDUs and edges represent the labelled relations between EDUs.  
%\cite{nishida2021outofdomain} 
The DDS will be used to help select relevant contexts for relation classification.  In the DDS, each directed edge (arrows) is a dependency.  The figure shows name of the relation as the text in red.  The direction of the relation indicates the importance of the information, with the arrowhead indicating the less important information.  By following the links back through the tree towards the root, one can select the relevant context for the classification task, which often is associated with information of greater importance.

\subsection{Context Selection Schemes}

A context selection scheme relies on two sub-steps: segmentation and filtering preceding text.  
%For segmentation, the aim is to divide the document into a sequence of EDUs, typically by first performing the more straightforward step of sentence segmentation.  
For this study, for a given EDU in the datasets, there are several options to select context.  The simplest filtering method is a \textit{null} method, where no context is used.  Alternatively, one can include a preceding text window of some size \textit{s}.   For example, we can also always select the previous sentence for a given argument.  However, just because some text precedes an argument does not necessarily mean it is relevant context that impacts text understanding.  Thus, we also explore a discourse-oriented method where a discourse structure is employed to identify other text (EDUs) that are linked via discourse relations (referred to here as \textit{structured context}).
%, usually impacting the interpretation of that text.  That is, without this context, one might interpret the argument text \textit{differently.}
%
This gives rise to the following schemes for context selection:

\paragraph{Default}  This baseline is a \textit{null} context (that is, no context is used).

\paragraph{(ADn) Add-n } This approach will add $n$ sentence that precede the first argument.

\paragraph{(ORn) Oracle-{n}} This scheme relies on the ground truth annotations to select the preceding context.\footnote{For our motivating example of \textit{contextualising claims}, in practice we would need an initial method to compute a discourse graph connecting EDUs.  We leave this to further work but note that prior work explores this task (e.g., \citeauthor{jeon-strube-2020-centering}, \citeyear{jeon-strube-2020-centering}).}  We use the following algorithm.  For a pair of arguments considered for DRC, we find the parent node of the \textit{first argument} as defined by the dependency discourse tree.
 
By chaining together the preceding context, in principle, we can vary the amount of context to include.  The intuition is that the discourse context, generally represented by a chain of EDUs following a path to the root, indicates which context is important enough to extract.

\section{Models}
\label{sec:models}

We focus on two general neural network language models approaches, both based on the Transformer architecture \cite{vaswani2023attentionneed}.  The first approach employs the RoBERTa model \cite{DBLP:journals/corr/abs-1907-11692}, an example of a non-auto-regressive (pre-trained) language model which one can finetune for classification.  The second approach uses large language models to perform prompt-based inference, typically used for generative AI.  
%Both have been shown to achieve state-of-the-art performance for various NLP tasks.
%in their respective classes.  

\subsection{PLM-finetuning with RoBERTa}

% \writingtodo{Rewrite this as a RoBERTa model}

Our RoBERTa-based approach is based on an approach that jointly models discourse connective generation and DRC  \cite{liu-strube-2023-annotation}.
% \textit{ConnRel} \cite{Liu2023a}.\footnote{https://github.com/liuwei1206/ConnRel}
This RoBERTa-based model performs training that combines two tasks: (1) the generation of a discourse connective that would link two arguments; and (2) the classification of the relation given the three pieces of information (argument 1, argument 2, and a connective).  As such, this model is extremely flexible and can be applied to both implicit and explicit DRC.
% \footnote{For more information, we direct the interested reader to the ConnRel documentation \cite{Liu2023a}.}  
%ConnRel uses the same RoBERTa network for both these tasks.  The network weights are updated during training using a combined linear loss for both tasks.  
% A related RoBERTa-based system was shown by the same authors to be highly competitive in relation classification in the DISRPT 2023 shared task \cite{Liu2023b} where the approach placed first \cite{Braud2023}.

%As an example of an approach using the transformer neural network architecture---the class of neural network responsible for some of the recent language modelling advances such as chatGPT---this represents an already comprehensive approach that utilises information within the data point, the benefits of the language modelling aspect of RoBERTa for generating a connective, and the benefits of the PLM for capturing useful linguistic "common sense" knowledge for discourse relation classification.

In this study, we generate variants of the data sets, subject to the preprocessing outlined in Section \ref{sec:data}, that differ by the amount and type of contextual information that is inserted \textit{before} the first argument (of the relation classification task).  That is, context added as per the selection schemes above.  The datasets are split into training and testing subsets to train and evaluate the RoBERTa model. 

We experimented with the full joint model described here and a simpler version that focuses just on classification.\footnote{This is achieved by setting the connective to a constant value for all data.}  The latter was found to perform the best and so we report only these results.  We used the default training setup and parameters, following the documentation in  \citet{Liu2023a}.\footnote{For computing environment, see Appendix A.}

\subsection{LLM-based Inference}

Prompt-based generative AI approaches using Large language models (LLMs) have been revolutionary in providing new baseline solutions for many tasks that apply across domains.  A key feature of LLMs are the comprehensive training regimes that potentially captures different kinds of knowledge, including common sense knowledge and linguistic capability (e.g., \citeauthor{DBLP:journals/corr/abs-2005-14165}, \citeyear{DBLP:journals/corr/abs-2005-14165}).
In this work, we examine two classes of LLMs for this inference approach: \textit{open} and \textit{closed} weight approaches.  For the former, we use Meta's LLaMA model \cite{grattafiori2024llama3herdmodels}.  For the latter, we use OpenAI's GPT4 \cite{openai2024gpt4technicalreport}.

\paragraph{LLaMA 3.1}
In this work, we use a locally hosted version of the LLaMa-3.1-70b-Instruct model, hosted on a server with 16 CPU cores, 128GB memory, and three A6000(48GB) which host the  Llama 3.1 70B model.

\paragraph{GPT4} 
For the GPT4 model, we use OpenAI's API with the \textit{chatcompletion} endpoint, using the gpt-4-0613 model.  

For both models,
%, this baseline is implemented with a cascaded approach.  However, for each component (connective generation and relation classification), 
for classification, we use In-Context Learning (ICL), widely acclaimed as having the ability to surpass supervised machine learning on many NLP tasks as a so-called ``few-shot learner'' \cite{DBLP:journals/corr/abs-2005-14165}.  This model helps us estimate how context impacts the current methods for LLM-based inference.  
%For these experiments, ICL using GPT is used for classification.
For this work, temperature was set to zero.

An example of prompt template for GPT4 is presented in Figure \ref{fig:prompt_template}. This frames the relation classification task as a MASK replacement generation task.  The prompt issues an instruction to replace the {[MASK]} token with one of the class labels provided in list format.  ICL "training" examples are provided, where we randomly sample one training data point from the data set for each class label.  The two arguments, the connective, and the {[MASK]} token are then added.  The Llama 3.1 prompt is similar.
%which comes after a pipe (``|'') delimiter.  The prompt template is 

\begin{figure}
\small
\texttt{Replace the MASK token (a discourse relation) by selecting only one of the following labels: [ \textit{label1}, \textit{label2}, ... , \textit{labeln}]  Examples: Passage 1: <\textit{arg$1_{ex1}$}>, Passage 2: <\textit{arg$2_{ex1}$}>, connective: <\textit{\text{connective}$_{ex1}$}> | \textit{label1} {EOL}}\\
\texttt{<further examples for remaining labels> {EOL}}  \\
\texttt{Passage 1: <\textit{arg1}>, Passage 2: <\textit{arg2}>, connective: <\textit{connective}> | [MASK]}
    \caption{GPT4 Prompt template for discourse relation classification}
    \label{fig:prompt_template}
\end{figure}

% \paragraph{Co-Star\footnote{As in the Kleene star}} In our variants of ConnRel, \textit{Contextual ConnRel, or CoCo-Rel}, and of GPT4.x, \textit{Co-GPT4.x}, we add varying amounts of context, drawn from the text immediately preceding the data point in question, using the full joint model of ConnRel which can then utilised this context in both learning to predict the connective and learning to predict the relation. We vary the amount of context based on the selection schemes outlined in the next subsection.

% For PDTB data, in a prior separate step, we use  ICL and prompt-based methods with the GPT4.x LLM to generate a connective.  This is necessary to avoid leaking \textit{expert manual annotations} for implicit connectives into the prompt, and provides a fair comparison against other DRC models, particularly for the implicit relations.  

\section{Experiment Results}
\label{sec:experiments}
We conduct an empirical study using ground truth linguistic data to examine the role of discourse structure in inferring discourse relations.  We deploy the described models in two conditions: a control condition without discourse context and an experiment condition that includes context. 

In the case of PLM-finetuning, for each pair of conditions, we run 10 different trials (that is, neural network training and testing with 10 different random seeds).  We report macro-F1 classification results averaged over the 10 trials.   For significance testing, we used the %Student's T-test (hereafter, T-test) \cite{student1908probable} for paired samples and the 
Wilcoxon Signed Ranked Test (WSRT) \cite{c4091bd3-d888-3152-8886-c284bf66a93a} and corrected for multiple comparisons using Bonferroni correction.
%\footnote{We test for significance with both parametric and non-parametric methods to minimise the chance of Type 1 and 2 statistical hypothesis testing errors.}   

We first examine if \textit{any} use of context leads to an improvement in the discourse classification task performance. For this investigation, we use n=1 for the context selection schemes.

\begin{table}[h]
\small
    \centering
    \begin{tabular}{|l|l|l|}
    \hline
     Approach    &  CovDTB & SciDTB         \\
    \hline
    \hline
    default      &   {75.54 (1.11)} &{57.42 (0.65)} \\

     AD1          &  73.45 (2.26) & \textbf{57.75 (1.08)}  \\

     \hline
     OR1          & {\textbf{75.78 (1.52)}}  &{\textbf{58.33 (0.77)}}$\dagger$  \\
     \hline 
    \end{tabular}
    \caption{Classification with a fine-tuned RoBERTa model. Macro-F1 scores (averaged over 10 runs) with standard deviations in parentheses.  Bolded values indicate improvements above the default. Daggers indicate statistical significance improvement using the Wilcoxon Signed Rank Test ($\dag:\alpha=0.05$).  %, $\ddag: \alpha=0.01$).     
    }
    \label{tab:Relation} 
\end{table}

\subsection{Context and PLM Fine-Tuning}

Table \ref{tab:Relation} shows the results of including context for the PLM-finetuning approach using the RoBERTa model. The table presents macro-F1 scores to give some indication of performance across the unbalanced dataset.\footnote{Here we report results for $n=1$ as our experiments showed that for larger values, adding more context confused the models.  Similarly, while we explored variants of the context representations that additionally utilised the relation class (for the linked context), the results were comparable to reported the OR1 variant, from which we conclude that the extra information did not help.}
We find that context generally helps for DRC when using a fine-tuned PLM, particularly when context is defined using discourse structure (OR1).  This improvement is statistically significant for the SciDTB dataset (WSRT $p<0.05$).
The AD1 context does not lead to strong performance improvements in comparison.  However, we note that the AD1 text window variant of context also helps mildly for the SciDTB but not for CovDTB.  

Across the two datasets, we observe that the performance results are higher in the CovDTB dataset compared to the SciDTB dataset.  This could be due to conventions in scientific writing for biomedical literature which may be more homogenous than the data from NLP domain found in SciDTB.

\subsection{Context and LLM Prompt-based Inference}

Table \ref{tab:Relation_model_GPT} presents the corresponding results for the GPT4 model.\footnote{Given the cost of using the commercial GPT LLM, we report results on single trials for the datasets.}  The results indicate a poor performance by the LLM for DRC under the default setting (with no context), even when using in-context learning.  Performance improves when discourse structure is used to provide context, as opposed to the adjacent text.  Indeed, performance drops when the adjacent sentence is used as context.
In the case of SciDTB, this brings performance closer to the PLM-finetuning result, within a margin of 5 F1 points.  However, while the DRC performance on CovDTB increases with discourse structure context, the macro-F1 scores still remain far behind the fine-tuned PLM, by a margin of over 25 F1 points.

In Table \ref{tab:Relation_model_llama}, we see a similar story, although Llama 3.1 performance is much lower than GPT4.  We suspect that this is primarily due to the size of the model; the Llama model used here has orders of magnitude fewer parameters than the GPT model.  Again, using the adjacent sentence leads to a drop in performance. Consistent with other results, an improvement in DRC performance was observed for SciDTB when using context defined by discourse structure.  Here, we failed to detect any improvement with the CovDTB dataset.  

\begin{table}[]
\small
    \centering
    \begin{tabular}{|l|l|l|}
    \hline
     Approach                     &  CovDTB & SciDTB  \\
    \hline
    \hline
    default      &     \multirow{1}{*}{32.07} &  \multirow{1}{*}{22.06 }\\
    
     AD1           &  \multirow{1}{*}{26.19} &  \multirow{1}{*}{19.28 } \\
    
     \hline
     OR1         &  \multirow{1}{*}{\textbf{49.07}}   &  \multirow{1}{*}{\textbf{52.61}} \\
    
     \hline 
    \end{tabular}
    \caption{GPT4 model: Classification macro-F1 scores.}
    \label{tab:Relation_model_GPT} 
\end{table}

\begin{table}[]
\small
    \centering
    \begin{tabular}{|l|l|l|}
    \hline
     Approach     &  CovDTB & SciDTB         \\
    \hline
    \hline
    default     &     \textbf{11.20} & 07.71  \\     
    AD1          &  10.04 &   05.15            \\
    
     \hline
     OR1         &  10.36 &  \textbf{11.15}  \\
     \hline 
    \end{tabular}
    \caption{Llama 3.1 model: Classification macro-F1 scores.}
    \label{tab:Relation_model_llama} 
\end{table}

We note that, while our focus is on comparison against our default baseline and the relative difference in performance with and without context, the models described in this paper are competitive with the reported performance in the literature.  We report these values for completeness in Table \ref{tab:sota}, which lists comparisons with literature, with the metrics generally reported by convention.  With the RoBERTa model and the ``oracle'' use of the ground truth discourse annotations used here, the measured accuracy of $83.63$ represents an estimate of an improvement over the prior state-of-the-art (SOTA) result for the CovDTB that could be obtained if we were to employ a fully automated version of the inference.

\begin{table}[h]
\small
    \centering
    \begin{tabular}{|l|l|l|}
    \hline
     Approach   &  CovDTB & SciDTB           \\
    \hline
    \hline
    Performance &  {70.03 }Acc. & {75.30} Acc.  \\
    \hline
    OR1 model   & \textbf{83.03} Acc. & 74.81 Acc.  \\
     \hline 
    \end{tabular}
    \caption{A comparison with performance reported in the literature.  \textbf{Bolded} values indicate where our best RoBERTa model surpasses previous results.  Reported performance  from: covdtb and scidtb \cite{Liu2023}.   \\
    }
    \label{tab:sota} 
\end{table}

To summarise, the results show positive trends for using discourse context for the DRC task. Generally, discourse context can help with the PLM-finetuning approach and LLM-inference.
When applying the text window context (AD1) results are mixed; the method does not work all the time and can decrease performance.  However, when using discourse structure to determine relevant context (OR1) we generally see improved performance, with stronger gains demonstrated with the SciDTB dataset.  This indicates that not all preceding text is useful for classification and that indiscriminately adding more context (without filtering) can make performance worse.

\subsection{A Reflection on Datasets}
\label{sec:discussion}

We speculate that one reason why we do not see a bigger effect from the inclusion of discourse context is that our datasets may be limited to relatively short length of the text data.  
Indeed, 
\citet{yang-li-2018-scidtb} note a related issue when studying news articles: discourse relations do not cross paragraph boundaries further making structures shallow.\footnote{This is presumably due to journalistic writing style.}  

In this regard, the CovDTB and SciDTB datasets, as examples of short text data (i.e., abstracts) may also have simpler discourse structures than longer texts.  We further investigated the nature of the discourse structures and found that, in the case of the SciDTB dataset, the structures were generally short-distance dependencies: 61\% of relationships are adjacent, with 10\% of relations separated by a gap of 3-5 sentences.  
We posit that when considering longer documents, the effect of structured context in DRC may be more pronounced.

\section{Error Analysis: DRC for Scientific Discourse Relations}
\label{sec:insights}

In the experiments presented above, we observed that providing context, particularly \textit{structured} context generally helps with DRC.  In this section, we perform an error analysis to better understand when context helps, analysing performance per relation type.  Here, we focus on the fine-tuned PLM approach as it yielded the highest macro-F1 scores.  For each of the 10 seed runs, we used predictions from the best models for the default (no context) and the OR1 (structured context) conditions.  Cases where the OR1 prediction was correct and the default was not was considered a \textit{win}.  The converse case was considered a \textit{loss}.  Where both approaches agreed, this was considered a \textit{tie}.  Margins for wins and loss were averaged over the 10 runs.

Table \ref{tab:positive_error_analysis} provides a list of the scientific discourse relation types in the \textit{winning} outcome for both data sets that benefited (overall) from OR1 context and their average win margins.  We can see that \textit{elaboration}, \textit{comparison}, \textit{attribution}, and \textit{temporal} relations were common to both datasets.

\begin{table}[h]
\small
    \centering
    \begin{tabular}{|l|l|} %  r|l|}
        \hline
         CovDTB & SciDTB \\
        \hline
        \textbf{elab}oration ($\Delta=5.7$) & \textbf{elab}-addition ($\Delta=1.5$)\\
          \underline{enablement} ($\Delta=1.2$) & \textbf{elab}-aspect ($\Delta=0.8$)\\
           cause-\underline{result} ($\Delta=0.8$)  & \textbf{temporal} ($\Delta=0.77$) \\
          \underline{condition} ($\Delta=0.4$) & \underline{bg}-\textbf{compare} ($\Delta=0.66$) \\
          \textbf{attribution} ($\Delta=0.4$) & joint ($\Delta=0.44$) \\
          \textbf{comparison} ($\Delta=0.3$)  & contrast ($\Delta=0.33$) \\
          \textbf{temporal} ($\Delta=0.1$) & progression  ($\Delta=0.22$) \\
          & exp-reason ($\Delta=0.22$) \\
        & \textbf{elab}-enum ($\Delta=0.22$) \\
        & \textbf{comparison} ($\Delta=0.11$) \\
        & \textbf{attribution} ($\Delta=0.11$) \\
    \hline
    \end{tabular}
    \caption{Winning relations: Discourse relations who DRC performance improved with the inclusion of structured context.   $\Delta=$ indicates the average win/loss margin.   Bolded text indicates potential correspondences across datasets.}
    \label{tab:positive_error_analysis}
\end{table}

Table \ref{tab:negative_error_analysis} presents the corresponding table for the \textit{losing} outcome.  Here we see that, both data sets have \textit{background} relations in common for this outcome.  If we assume \textit{findings} and \textit{results} are related relations, then we can consider this a further potential alignment.
    
\begin{table}[h]
\small
    \centering
    \begin{tabular}{|l|l|} %  r|l|}
         \hline
         CovDTB & SciDTB \\
        \hline
        \textbf{findings} ($\Delta=0.9$) &  \textbf{bg}-goal ($\Delta=0.44$)\\
        \textbf{background} ($\Delta=0.3$) & manner-means ($\Delta=0.44$)\\
        & \underline{enablement} ($\Delta=0.44$) \\
        & evaluation ($\Delta=0.22$) \\
        & \textbf{result} ($\Delta=0.22$) \\
        & \textbf{bg}-general ($\Delta=0.22$) \\
        & \underline{condition} ($\Delta=0.11$) \\
        & exp-evidence ($\Delta=0.11$) \\
    \hline
    \end{tabular}
    \caption{Losing relations: Discourse relations who DRC performance suffered with the inclusion of structured context.  $\Delta=$ indicates the average win/loss margin.  Bolded text indicates potential correspondences across datasets.}
    \label{tab:negative_error_analysis}
\end{table}

Table \ref{tab:neutral_error_analysis} shows the relations that had an equal number of wins and losses.  We present these for completeness.  However, it may be the case that, for these datasets, there is insufficient data to assign these to either the winning or losing outcomes.

\begin{table}[h]
\small
    \centering
    \begin{tabular}{|l|l|} %  r|l|}
        \hline
         CovDTB & SciDTB \\
        \hline
          textual-organisation & elab-definition \\
          manner-means & elab-process-step \\
           & cause \\
    \hline
    \end{tabular}
    \caption{Tied relations: Discourse relations who DRC performance remained the same with the inclusion of structured context.}
    \label{tab:neutral_error_analysis}
\end{table}

There were some differences between datasets for a subset of relations, which were placed in different outcomes (winning, losing).  These included \textit{enablement} and \textit{condition}.  In CovDTB, a single relation is used for \textit{cause-result} which was in the winning outcome.  For the SciDTB dataset, the \textit{result} relation was in the losing outcome.  Similarly, while most background relations were in the losing outcome for both datasets, \textit{bg-compare} was in the winning outcome for SciDTB; though this could be because the winning outcome contained more comparison-related relations.  We treat these divergences as interesting outcomes to investigate further, noting that some of these may be due to annotation differences between the datasets.

In Table \ref{tab:analysis_examples}, we present some examples of data as assigned to the winning and losing outcomes.  For the winning outcomes, the high-level statement of the research activity as context may contribute positively to the DRC task.  For the losing outcome, we note that in the SciDTB example, the high-level context may simply be too broad.  For the CovDTB, we note that both findings and background relations tended to be at the beginning of the text and so no prior context exists, explaining why these relations are in the losing outcome.

\begin{table*}[t]
\small
    \centering
    \begin{tabular}{|c|c|c|l|}
    \hline
    Condition & Dataset & Relation     &  Example\\
    \hline
    \hline
    %wan049@virga-login:~/itch3/work/jca2023/hits-home/home/wansn/work/code/stephen-wan/ConnRel/data/dataset/eng_dep_scidtb/withContext/mode-1-context-1/eng_dep_scidtb/fine/preds> grep 'comparison' less joint+test_l1+7+106464.txt 
    winning & scidtb & comparison     & \textbf{Context}: We propose a novel method of jointly embedding entities and words into \\
    & & & the same continuous vector space . \\
    & & & \textbf{Arg1}: that jointly embedding brings promising improvement in the accuracy\\
    & & & of predicting facts ,\\
    & & & \textbf{Arg2}: compared to separately embedding knowledge graphs and text .       \\
    \hline
    %wan049@virga-login:~/itch3/work/jca2023/hits-home/home/wansn/work/code/stephen-wan/ConnRel/data/dataset/eng_dep_scidtb/withContext/mode-1-context-1/eng_dep_scidtb/fine/preds> grep 'result'  joint+test_l1+7+106464.txt 

    losing & scidtb & result     & \textbf{Context}: We describe a search algorithm  \\
    & & & \textbf{Arg1}: Our results show\\
    & & & \textbf{Arg2}: parsing results significantly improve       \\
    \hline
    %wan049@virga-login:~/itch3/work/jca2023/hits-home/home/wansn/work/code/stephen-wan/ConnRel/data/dataset/eng_dep_covdtb/withContext/mode-1-context-1/eng_dep_covdtb/fine/preds> grep 'comparison'  joint+test_l1+10+106524.txt 
    winning & covdtb & comparison     & \textbf{Context}: Herein we discuss application of the Collaborative Cross ( CC ) panel of \\
    & & & recombinant inbred strains  \\
    & & & \textbf{Arg1}: Although the focus of this chapter is on viral pathogenesis ,\\
    & & & \textbf{Arg2}: many of the methods are applicable to studies of other pathogens ,\\
    & & & as well as to case-control designs in genetically diverse populations .\\
    \hline
    %wan049@virga-login:~/itch3/work/jca2023/hits-home/home/wansn/work/code/stephen-wan/ConnRel/data/dataset/eng_dep_covdtb/withContext/mode-1-context-1/eng_dep_covdtb/fine/preds> grep 'findings'  joint+test_l1+8+105202.txt 
    losing & covdtb & findings     & \textbf{Context}: ROOT (no context)
    \\
    & & & \textbf{Arg1}: In this work , we demonstrate a design of meta - holography  \\
    & & & \textbf{Arg2}: that can achieve 2 28 different holographic frames and an extremely high \\
    & & & frame rate in the visible range . \\
    \hline
    \end{tabular}
    \caption{Examples of discourse relations in the winning and losing outcomes for both the SciDTB and CovDTB datasets.}
    \label{tab:analysis_examples}
\end{table*}

To dive deeper into what might potentially explain the difference between winning and losing outcomes, we examined the first word of the second argument, checking for a match against a list of known discourse connectives.\footnote{This list was based connectives from PDTB data sets (2 and 3) and the collated  connectives from the DiscoGEM dataset \cite{scholman2022DiscoGeM}. URL: https://github.com/merelscholman/DiscoGeM/tree/main/Appendix}  Here, we make the simplifying assumption that the discourse connective is found between the two arguments.

Table \ref{tab:connective_matches} shows the percentage of instances where, for relations in either the winning or losing outcome, the first word of the second argument was a known discourse connective.
We observe that, for both datasets, winning outcomes exhibit a higher percentage of matches for connectives.
%, though the difference is larger CovDTB than SciDTB.  
We take the matches as a potential indicator of the higher proportion of explicitly marked discourse relations.  This raises the potential hypothesis that perhaps context may be more beneficial for DRC of certain explicitly marked relations.  
%The larger difference noted in the case of CovDTB could be why the performance was starker in performing DRC with context for this dataset.

\begin{table}[t]
\small
    \centering
    \begin{tabular}{|r|c|c|}
    \hline
    \multirow{2}{*}{Relation Category} & \multicolumn{2}{|c|}{Percentage of connective matches} \\
     & CovDTB & SciDTB \\
    \hline
    Losing Relations &$7.8\%$ &$25.0\%$ \\
    Winning Relations & $\textbf{16.6\%}$ & $\textbf{28.2\%}$\\
    \hline

    \hline
    \end{tabular}
    \caption{Percentage of matches to a list of explicit connectives across the positive, neutral and negative relations.}
    \label{tab:connective_matches}
\end{table}

\section{Future Work}
\label{sec:future}

Our preliminary investigation here on the role of context for DRC in scientific writing highlights two potential avenues for future research.  Our error analysis suggests that structured context may potentially be more beneficial the DRC for certain explicit relations (for scientific writing).  We intended to further investigate this.

We note that our investigation here is limited to dependency discourse structures and the representation of context as string concatenations.  In subsequent work, we aim to explore different automatically inferred graph representations of text structure, particularly longer text documents.
%, noting that focusing on directed acyclic graphs more generally may be of interest.  

Our experiments were also limited to two categories of LLM-based inference, namely In-Context Learning (ICL) for closed and open weight LLMs (or proprietary and so-called "open-source" LLMs).  In the future, we intend to include LLMs that include some reasoning capability, such as the recent GPT-o1 and DeepSeek models, as well as techniques like chain of thought, to see if these inference methods help with DRC.  In this work, we also used one example of a transformer network for PLM fine-tuning.  In future work, we aim to experiment with the model of \cite{gessler-etal-2021-discodisco} as an alternative competing transformer model.

Finally, returning to our motivating example, we intend to examine the role of discourse relations in identifying relevant supporting source material to validate generative AI output.  We intend to conduct qualitative and quantitative user studies to better understand the potential for discourse information to help with these goals.
\section{Conclusions}
\label{sec:conclusion}

In this work, we showed that adding discourse context, particularly \textit{structured context}, helps with Discourse Relation Classification for scientific writing.  We demonstrated this using two dominant neural language modelling methods: finetuning using a pre-trained language model, and inference with large language models using in-context learning.  The analysis presented here focuses on two scientific discourse datasets, CovDTB and SciDTB, representing biomedical and computer science disciplines.  We found that, for the science discourse relations represented in these datasets, context might help for specific relations, such as with \textit{elaboration}, \textit{attribution}, \textit{comparison} and \textit{temporal} relations.

\section*{Acknowledgements}
This work was funded by the CSIRO Julius Career Award.  We are also grateful to the Heidelberg Institute for Theoretical Studies (H-ITS) for supporting this project and providing facilities for conducting this research.  We would like to further acknowledge the the feedback from the CSIRO Language Technology team, the H-ITS NLP team, and the anonymous reviewers on previous versions of this paper.

\section*{Limitations}
In this work, we focused on English prose language data from publicly available datasets. As such, our conclusions about discourse relations, connectives and the need for using context for discourse relation classification are limited to this language and the genres represented.  We note that while we are interested in scientific writing in general, here we study data from just two science disciplines: computer science and biomedical articles about Covid.  We note that we only generated a single set of results using prompt-based methods (with Llama 3.1 and GPT 4), using a temperature of zero, due to costs.  It is possible that multiple trials of the approach may yield different results.  Finally, we note that prompt engineering was limited.  It may be possible that stronger performance gains may be obtained if further prompt engineering is employed.  For further limitations, see our future work section.

\section*{Ethical Considerations}
In this work, we use publicly available discourse-related datasets.  Our analysis is focused discourse-related linguistic phenomena and is not focused on any individual or subgroup in the community.  The work, while motivated by current trends in applied AI, is not immediately applicable in real-world usage.

% wei added the "custom" below
\bibliography{manual}
\bibliographystyle{acl_natbib}

\appendix

\section{Appendix: Computing Environment}
\label{sec:appendix}

Experiments for training and evaluating the ConnRel model (RoBERTa-based, 82M parameters) were conducted on a server with 1 node (4x NVIDIA A40; 2x Intel(R) Xeon(R) Gold 6330 CPU @ 2.00GHz; 32GB RAM).  Each of the 3 approaches tested was trained and evaluated with 5 datasets, over  10 trials. Each trial ranged from between 30 minutes to 1.5 hours, depending on the dataset.  Estimated GPU time per approach is 36 hours.  Experiments were also repeated at least twice to test for replicability. This results in  approximately, 432 hours of GPU time (single jobs).

% This is an appendix.
% \input{appendices}
% \input{old_tables}

\end{document}